\documentclass[journal]{IEEEtran}

\usepackage{lineno,hyperref}
\usepackage{listings}
\usepackage[]{algorithm2e}

\ifCLASSINFOpdf
   \usepackage[pdftex]{graphicx}
\else
\fi

\usepackage{url}
\begin{document}
\title{Kafka-ML: connecting the data stream with ML/AI frameworks}

\author{Cristian~Mart\'in, Peter Langendoerfer, Pouya~Soltani~Zarrin, Manuel~D\'iaz and Bartolom\'e~Rubio
\thanks{Cristian~Mart\'in, Manuel~D\'iaz and Bartolom\'e~Rubio are with the ITIS software and the Department of Languages and Computer Science at the University of M\'alaga, M\'alaga, Spain. %Cristian~Mart\'in and Peter Langendoerfer are with IHP - Leibniz-Institut für Innovative Mikroelektronik, Frankfurt (Oder), Germany. Peter Langendoerfer is also with BTU Cottbus-Senftenberg, Cottbus, Germany. 
Peter Langendoerfer and Pouya~Soltani~Zarrin are with IHP - Leibniz-Institut für Innovative Mikroelektronik, Frankfurt (Oder), Germany. Peter Langendoerfer is also with Brandenburg University of Technology Cottbus-Senftenberg, Cottbus, Germany.
E-mail: Cristian Mart\'in (cmf@lcc.uma.es).}}


\markboth{Cristian Mart\'in \MakeLowercase{\textit{et al.}}: Kafka-ML: connecting the data stream with ML/AI frameworks, ArXiv, June~2020}%
{Shell \MakeLowercase{\textit{et al.}}: Bare Demo of IEEEtran.cls for IEEE Journals}

\maketitle

\begin{abstract}
Machine Learning (ML) and Artificial Intelligence (AI) have a dependency on data sources to train, improve and make predictions through their algorithms. With the digital revolution and current paradigms like the Internet of Things, this information is turning from static data into continuous data streams. However, most of the ML/AI frameworks used nowadays are not fully prepared for this revolution. In this paper, we proposed Kafka-ML, an open-source framework that enables the management of TensorFlow ML/AI pipelines through data streams (Apache Kafka). Kafka-ML provides an accessible and user-friendly Web User Interface where users can easily define ML models, to then train, evaluate and deploy them for inference. Kafka-ML itself and its deployed components are fully managed through containerization technologies, which ensure its portability and easy distribution and other features such as fault-tolerance and high availability. Finally, a novel approach has been introduced to manage and reuse data streams, which may lead to the (no) utilization of data storage and file systems.
\end{abstract}

\begin{IEEEkeywords}
Kafka-ML, Apache Kafka, Machine Learning, Artificial Intelligence, Data Streams, TensorFlow, Docker, Kubernetes
\end{IEEEkeywords}

\IEEEpeerreviewmaketitle

\section{Introduction}
In this digital era, information is continuously acquired and processed everywhere, from many sources and for many purposes and sectors. In this sense, Machine Learning (ML) and Artificial Intelligence (AI) \cite{lu2019artificial} are playing a decisive role in converting raw information into useful predictions and recommendations to improve both business operations and the life of citizens, among other. For instance, companies like Facebook process millions of photos every day to detect inappropriate content. This creates a continuous \textit{data stream} of information that is facing ML/AI algorithms and systems.

More recently, with the proliferation of the Internet of Things (IoT) \cite{diaz2016state}, new sources of data have been enabled in the Internet era, with a forecast of 500 billion of connected devices by 2030 \cite{cisco2030}.  Paradigms such as Industry 4.0, Connected Cars and Smart Cities are being possible, and the most important, they contribute to the digitization of services and phenomena of the physical world. As a result, the data stream has  continuously been increased and forecasts predict a huge expansion for coming years.

Traditionally, most of the ML/AI frameworks, which are behind the design and development of ML/AI algorithms, have not been designed to work with data streams, but with persistent datasets and static data. Even nowadays, popular Python frameworks like PyTorch, Theano and TensorFlow do not provide or only provide partial support for data stream systems like Apache Kafka \cite{kafka}, the most popular data stream system. This does not only include training of ML models, but also the rest of the steps that may be part of an ML/AI pipeline, such as inference, testing, and evaluation. To cope with this challenge, Kafka-ML\footnote{https://github.com/ertis-research/kafka-ml}, an open-source framework to manage ML/AI pipelines through data streams is presented. Kafka-ML makes use of Apache Kafka and currently supports TensorFlow as ML framework to integrate data streams and ML/AI. However, the goal is to extend the support for ML/AI frameworks in the near future. Kafka-ML offers an accessible and user-friendly Web User Interface (following a similar approach as AutoML initiatives) to manage the ML/AI pipeline for both experts and non-experts on ML/AI. Users just need to write a few lines of ML model code to train, compare, evaluate and do inference on their algorithms. Moreover, this framework makes use of a novel approach to manage data streams in Apache Kafka, that can be reused as many times as they are configured leading to the (no) need for any data storage or file system for datasets in Kafka-ML.  Finally, Kafka-ML exploits containerization and container orchestration platforms to distribute the load of the system and facilitating the distribution of its components, in addition to providing fault-tolerance and high availability.

Therefore, the main contributions of this paper are:
\begin{enumerate}
    \item The presentation of Kafka-ML, an open-source, accessible and user-friendly framework to manage ML/AI pipelines through data streams
    
    \item A novel approach to manage the data streams of ML/AI pipelines with (no) need for data storage or file systems
\end{enumerate}

The rest of the paper is organized as follows. Section \ref{sec:background} presents a background of Kafka-ML.  Section \ref{sec:motivation} presents the motivation of this work. In Section \ref{sec:pipeline} the ML/AI pipeline of Kafka-ML is introduced. Then, in Section \ref{sec:architecture} the Kafka-ML architecture and its components are presented. The approach for data stream management in Apache Kafka is presented in Section \ref{sec:storage-kafka}. A validation of Kafka-ML is analyzed in Section \ref{sec:validation} and related work is discussed in Section \ref{sec:related-work}. Lastly, our conclusions and future work are presented in Section \ref{sec:conclusions-future}.

\section{Background}
\label{sec:background}

Apache Kafka is a distributed messaging system (publish/subscribe) that can dispatch and consume large amounts of data at low latency. Traditional message queues can support high rates of message consumption by adding multiple consumers per topic, but only one consumer will receive each message at a time. Like message queues, publish/subscribe systems exchange information from producers to consumers. Nevertheless, in contrast to message queues, publish-subscribe systems allow  multiple consumers to receive each message in a topic. Nowadays in the era of big data, stream data goes to multiple systems like batch processing and stream processing, but also a low latency is required. Therefore, both features are required, and this is how Apache Kafka provides them:

\begin{itemize}
    \item Multi-customer distribution. As a publish/subscribe system, Apache Kafka enables the connection of multiple clients and customers to messages. Moreover, thanks to its integration and support for a wide range of solutions like Apache Hadoop, Apache Storm, TensorFlow, etc., this feature is definitely more than possible. 
    
    \item High rate of message dispatching. This is achieved by a conjunction of functionalities: 1) message set abstractions:  messages are grouped together amortizing the overhead of the network round trip rather than sending a single message at a time; 2) binary message format: data chunks can be transferred without modifications; and 3) zero-copy optimizations: to avoid many copies of the pagecache. However, one of the most notable features is the Kafka consumer group, which enables the distribution of messages in a cluster of customers managed by Apache Kafka like message queues.
\end{itemize}

Topics are the stream of messages in Kafka, wherein producers can publish messages and consumers can subscribe to receive them. When a message is sent by a producer to Kafka, on the contrary to many distributed queue frameworks, Kafka stores it in disk with a configurable retention policy, enabling later data retrieving by components. This is popularly known as the \textit{distributed log}, and enables consumers to go through the log as they have to. In some cases, like ML training in Kafka-ML, this feature is suitable since all data is processed at once and whether a failure occurs during this process the customer can start again without losing any data and having to store it in a file system. 

Load balancing and fault-tolerance are also performed by partitions of the topics, where each topic can be divided into multiple partitions, and each partition can have multiple replicas. Partition enables the log to be divided into smaller units and providing load balancing, and the topic replicas enable fault-tolerance. An Apache Kafka cluster is composed of a peer-to-peer network of Brokers that share partitions and replicas. When having a consumer group, partitions can be associated to customers enabling high dispatching rates. Apache Kafka also incorporates different policies such as 'at most one', 'at least once' and 'exactly one', which enables customized QoS policies for message dispatching.

Its popularity, its large number of implementations and integrations with many cloud computing systems, and its great acceptance in the community have converted Apache Kafka into the de facto solution for interconnecting systems, ingesting data and dispatching information.

\section{Motivation}
\label{sec:motivation}

It is indubitable that ML/AI applications require a bunch of data, sometimes large amounts when having video data sources, to train and evaluate their algorithms. Traditionally, ML/AI data is found at data stores and large files, which difficult the sharing between users and applications. As a result, in most cases, users and developers have a local copy of the dataset (let's suppose the impact of this for a medium-big company) or use shared infrastructures to process and access datasets. On the other hand, the same datasets may be needed to evaluate different algorithms and approaches, which may be limited in systems not designed for dispatching data.

With the proliferation of the IoT, data is turning from static sources to data streams. The Internet is another huge source of information and data streams. However, most of current ML frameworks have not been designed to work with data streams, i.e., they directly do not support them or they just provide a connector but have not been properly integrated with ML/AI pipelines. This does not only involve inference tasks where data streams can be used to make predictions for incoming data, but also for the ingestion of data streams for the training and evaluation phases of ML/AI pipelines. Working and having with data streams along network links can incur in data loss, which may not be present in local file systems and should also be considered. 

Furthermore, we aim at providing ML and AI open and accessible for everyone. In this sense, AutoML initiatives such as Google Cloud AutoML have contributed to bring high-quality models and solutions, adapted to multiple business needs, to developers with limited experience. However, these initiatives have not been yet properly integrated with data stream pipelines. Another way to pave the way of users into ML and AI is providing a missing ecosystem where they can share trained models and metrics (e.g., loss and accuracy) that can also be used to evaluate different models and configurations.

ML/AI solutions usually need requirements that are difficult to be accomplished in personal computers, and developers tend to adopt shared infrastructures to deploy their applications. Moreover, high-availability, load-balancing and fault-tolerance may be required in ML/AI mission-critical applications and should be provided in a transparent way to users. This would enable users and developers to focus on ML models and the business logic, facilitating the life cycle of ML/AI applications in production systems.  

To sum up, we envisage a system that reduces the gap between data streams and ML frameworks, adopting AutoML features and enabling the sharing of ML models, metrics and results in high-performance and high-availability infrastructures.

\section{Pipeline of an ML model in Kafka-ML}
\label{sec:pipeline}

\begin{figure}
    \centering
    \includegraphics [scale=0.3]{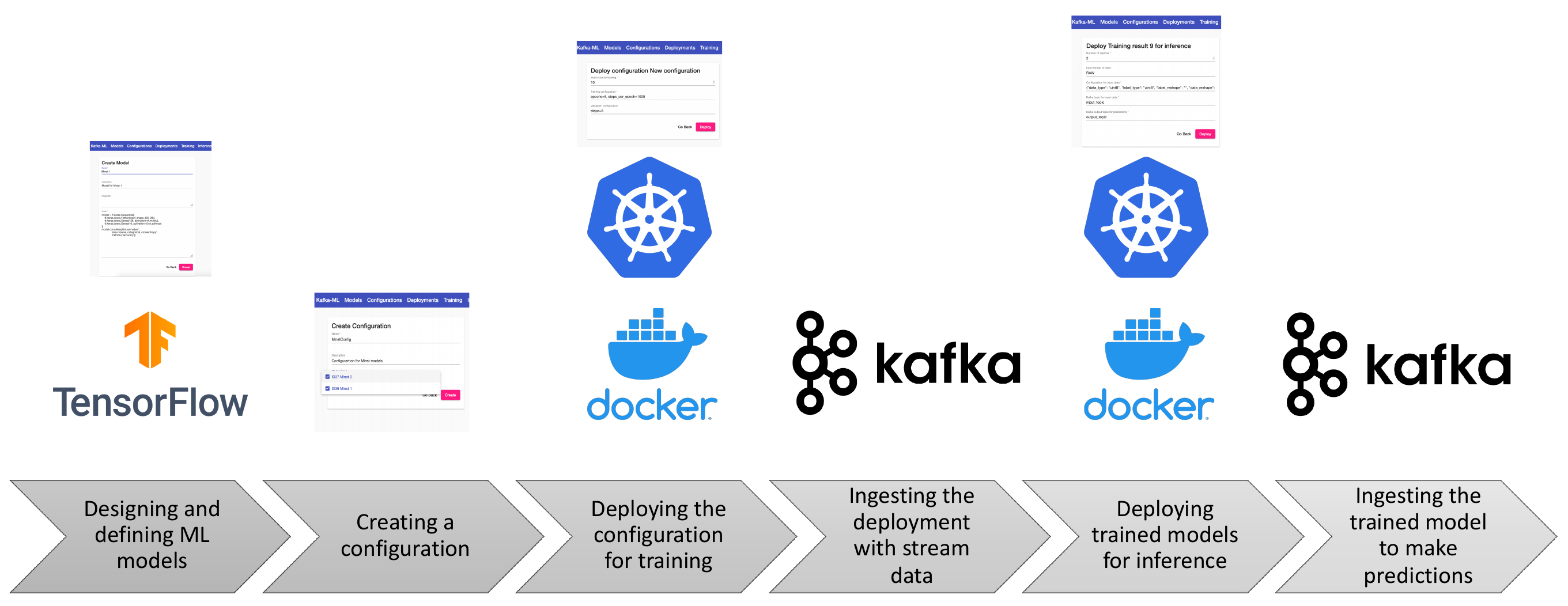}
    \caption{ML/AI pipeline in Kafka-ML}
    \label{fig:pipeline}
\end{figure}

In this section, we introduce the pipeline of an ML model in Kafka-ML representing its life cycle. Fig. \ref{fig:pipeline} depicts the pipeline and steps to be carried out: A) designing and defining the ML model; B) creating a training configuration of ML models, i.e., selecting a set of ML model(s) to be trained; C) deploying the configuration for training; D) ingesting the deployed configuration with training and optionally evaluation stream data through Apache Kafka; %E) visualizing the training results of the ML model(s) and decide which model(s) should be deployed for inference; 
%E) and finally, deploying the trained model for inference, which can be fed to make predictions with data streams. 
E) deploying the trained model for inference; F) and finally, feeding the deployed trained model for inference to make predictions with data streams. Since most of previous steps use a RESTful API, the pipeline can be automatized, and all the steps related to feeding the ML model (training and inference) are carried out with data streams. Datastores might be not needed anymore (Section \ref{sec:storage-kafka}). In the following, each of such steps is detailed.

\subsection{Designing and defining ML models}
From the first moment, we wanted to make this step as simple as possible to let ML developers focus on ML models instead of learning a new library or using complex pipelines. A tool that can enable easy testing and validation of ML models for ML developers would considerably facilitate their work and would let them focus on what they are experts on. For this reason, the only source code needed is the ML model definition itself in a popular ML framework as shown in Listing \ref{list:definition}.
\
\begin{lstlisting}[caption={Example ML code for Kafka-ML}, label={list:definition}]
model = tf.keras.Sequential([
    tf.keras.layers.Dense(32,input_dim=100 
        ,activation=tf.nn.relu),
    tf.keras.layers.Dense(1, 
        activation=tf.nn.sigmoid)])
model.compile(optimizer=`adam',
    loss=`sparse_categorical_crossentropy'
    metrics=[`accuracy'])
\end{lstlisting}              

Listing \ref{list:definition} source code may seem familiar. In fact, it is a simple Python TensorFlow/Keras model with a hidden layer, a single output and the compilation for training. Kafka-ML currently supports Python TensorFlow \cite{abadi2016tensorflow} due its support for Apache Kafka through TensorFlow/IO\footnote{https://www.tensorflow.org/io}. Currently, Apache Kafka integrations are under development and its Kafka-ML domain is getting expanded by receiving further ML frameworks.

\begin{figure}
    \centering
    \includegraphics [scale=0.22]{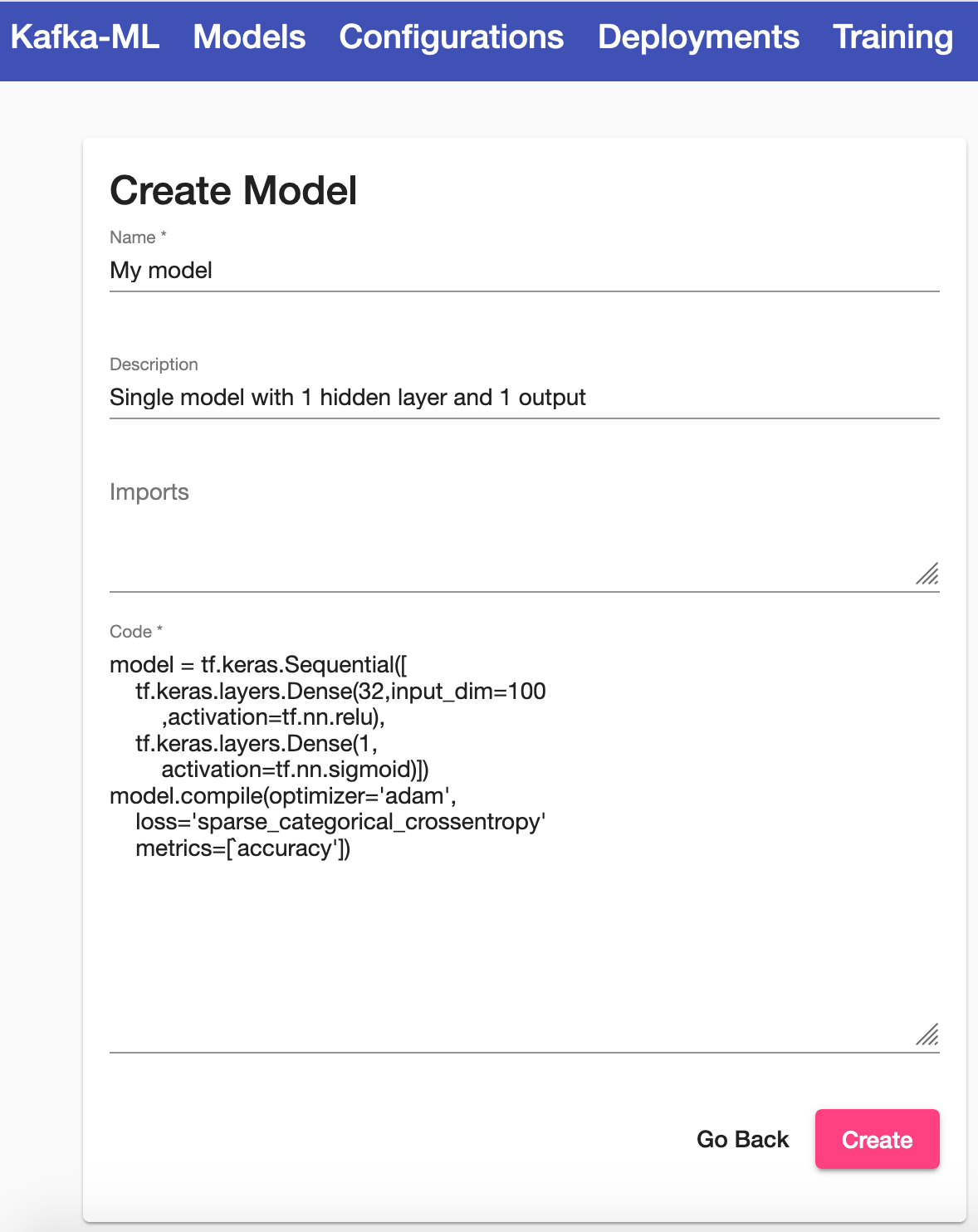}
    \caption{Definition of an ML model in Kafka-ML}
    \label{fig:create-model}
\end{figure}

Once a model is defined using a ML editor (e.g, Jupyter), the TensorFlow/Keras source code of the model can be inserted into the Kafka-ML Web UI (User Interface) for model creation as shown in Figure \ref{fig:create-model}. Note that the model can also be defined directly on Kafka-ML, though it is recommended to validate it beforehand using other and more powerful ML  Integrated Development Environment (IDE) or editors. Other required functions for the model (if any) can be inserted in the \textit{imports} field. Once the model is submitted, the source code will be checked as a valid TensorFlow model and incorporated into Kafka-ML. If the model has been successfully defined, the pipeline can be continued to the next step.

\subsection{Creating a configuration}

\begin{figure}[h]
    \centering
    \includegraphics [scale=0.3]{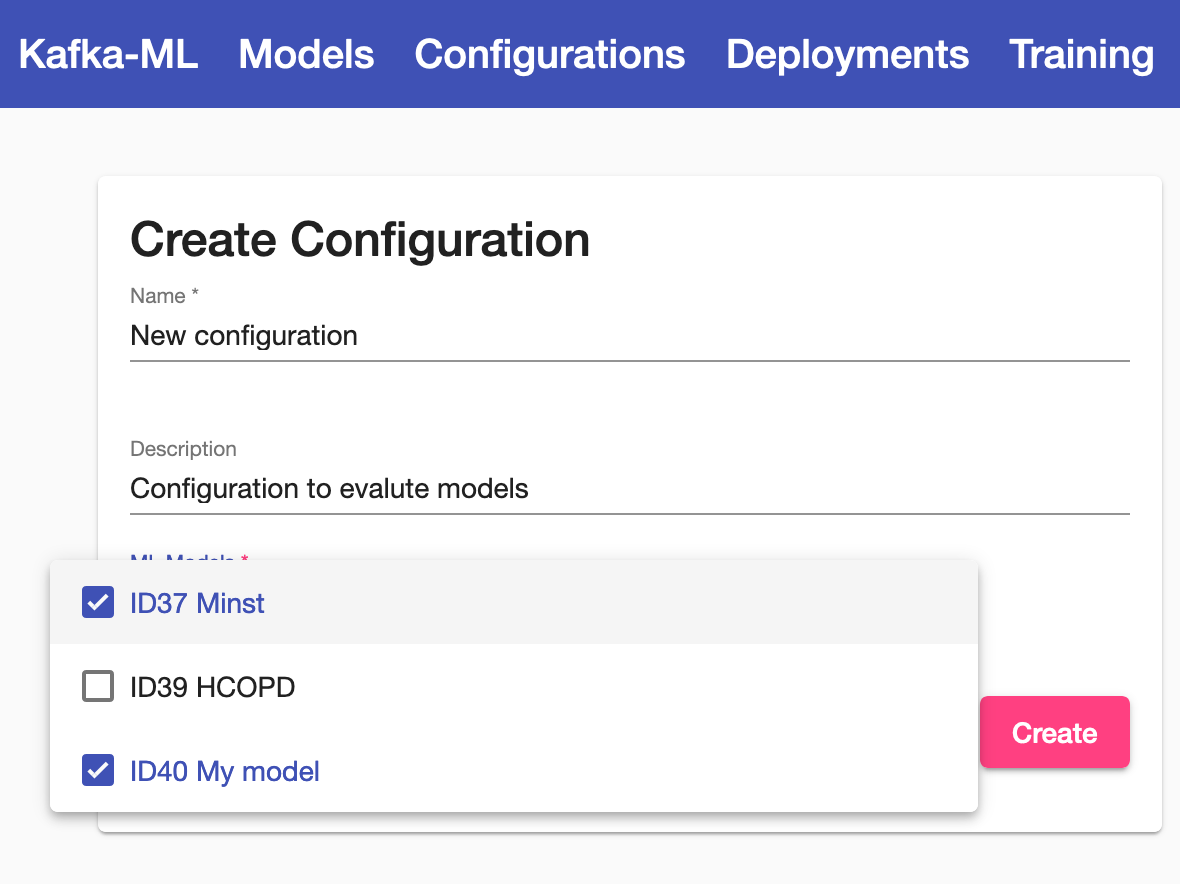}
    \caption{Creation of a configuration in Kafka-ML}
    \label{fig:create-configuration}
\end{figure}

A configuration is a logical set of Kafka-ML models that can be grouped for training. This can be useful when it is required to evaluate and compare metrics (e.g., loss and accuracy) of a set of Kafka-ML models or just to define a group of them that can be trained with the \textit{same} and \textit{unique} data stream in parallel. Therefore, in case of having $n$ ML models, which all of them require a data stream for training, just only one data stream has to be sent to Apache Kafka if a configuration has been defined with the $n$ models. Note that a configuration can also be defined with only a model. A configuration can be created in the Kafka-ML Web UI as shown in Fig. \ref{fig:create-configuration}.

\subsection{Deploying the configuration for training}

\begin{figure}[h]
    \centering
    \includegraphics [scale=0.30]{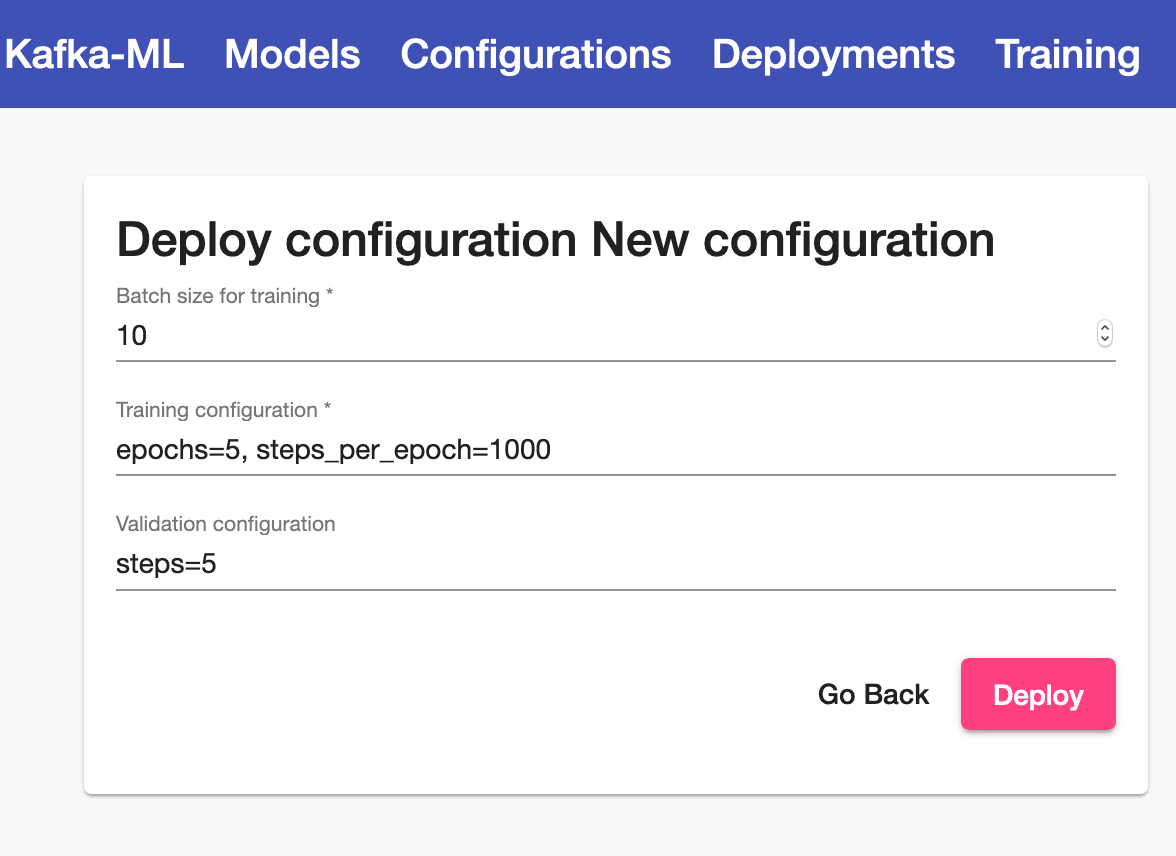}
    \caption{Deploying a configuration for training in Kafka-ML}
    \label{fig:deploy-configuration}
\end{figure}

After setting some training parameters like the batch size, epochs and number of iterations, and optionally some parameters for evaluation in the Kafka-ML Web UI (Fig. \ref{fig:deploy-configuration}), the configuration will be ready to be deployed for training. If so, a task will be deployed per Kafka-ML model. Then, one of the first steps that each deployed job carries out is fetching its corresponding ML model from the Kafka-ML architecture and loading it to start training. Finally, jobs can resume until a data stream with training and optionally evaluation data is received through Apache Kafka. This allows both having ready-to-train ML models when a data stream is sent and direct training if the data stream is already in Kafka.

\subsection{Ingesting the deployment with stream data}

Once the models are deployed, for the continuation of the pipeline, the data stream has to be sent for training. It can also be submitted before the model deployment. Since the Kafka stream connector expects to have the stream data at the initiation, the training cannot start until the data stream is available in Kafka. We have used at least two Kafka topics to overcome this: 1) data topic(s) which only contain training and evaluation data streams required for training and evaluation; 2) and a control topic, which informs deployed ML models through control messages \textit{when} and \textit{where} the data streams are available for training and evaluation. Section \ref{sec:storage-kafka} will discuss this in detail. A control message should contain at least the following information:

\begin{figure*}[h]
    \centering
    \includegraphics [scale=0.3]{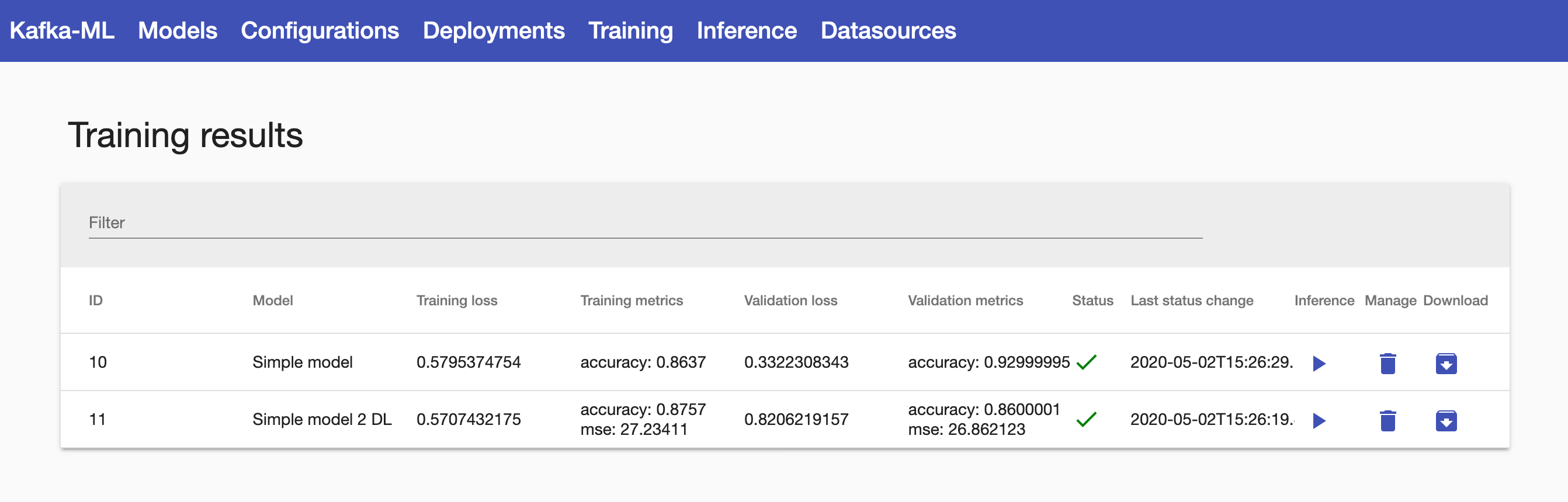}
    \caption{Training management and visualization in Kafka-ML}
    \label{fig:training-results}
\end{figure*}

\begin{itemize}
    \item \textbf{deployment\_id}: ID of the deployed configuration where the data stream goes.
    \item \textbf{topic}: Kafka topic where the training and evaluation data streams are.
    \item \textbf{input\_format}: Format of the data stream (e.g, RAW, AVRO).
    \item \textbf{input\_config}: Configuration required by the data format chosen (e.g., the scheme in Avro).
    \item \textbf{validation\_rate}: Percentage of stream data that will be used for evaluation. If validation\_rate is equal to zero, only training will be performed. 
    \item \textbf{total\_msg}: Number of messages dispatched in the data stream. The number of messages sent are calculated by the Kafka-ML libraries.
\end{itemize}

Kafka-ML currently supports RAW format (suitable for single-input data streams that may request a reshape, like images) and Apache Avro (suitable for complex and multi-input datasets where a scheme specifies how the data stream is decoded), however, it is open for the support of new data formats. In each case, the information for decoding is included in the control message (input\_config), as for example, the training and label data schemes for the Avro format. We have developed libraries for these two data formats, which make the data stream dispatching easier since they deal with Kafka-ML aspects like sending the control message when the data stream has been sent. 

After dispatching the data stream with the libraries provided from an IoT device or gateway, a dataset or any information source to the corresponding deployment\_id, all the ML models grouped in the configuration will start the training, and evaluation (if $validation\_rate>0$).

\subsection{Deploying trained models for inference}
Right after training and evaluation, both the trained model itself and its defined metrics (e.g., loss and accuracy) will be submitted by each training Job to the Kafka-ML architecture. Results can be visualized in the Kafka-ML Web UI as shown in Figure \ref{fig:training-results}. For each result, users can edit it, download the trained model, or deploy it for inference.

\begin{figure}
    \centering
    \includegraphics [scale=0.3]{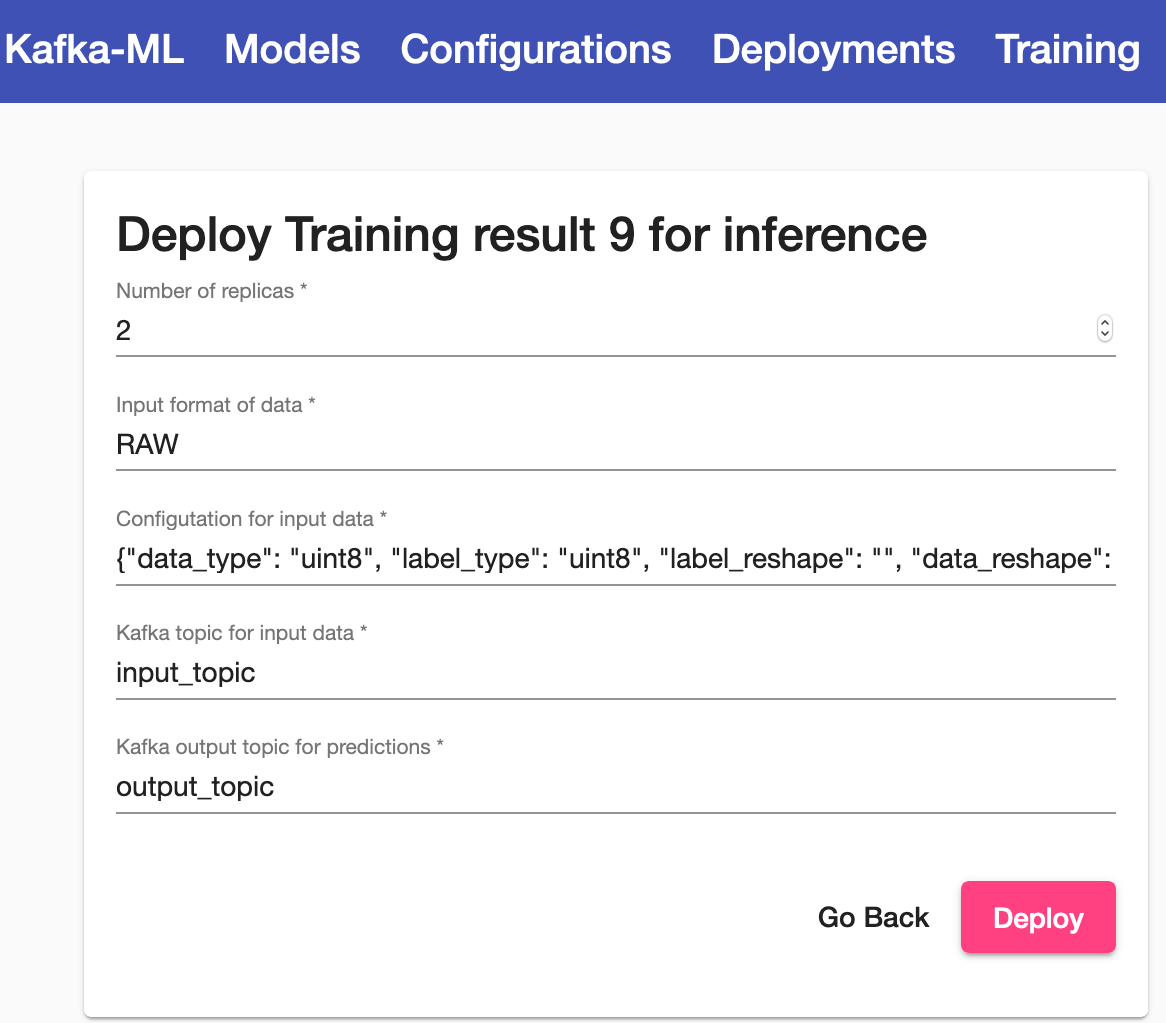}
    \caption{Deploying a trained ML model for inference in Kafka-ML}
    \label{fig:deploy-inference}
\end{figure}

In the inference deployment (Fig. \ref{fig:deploy-inference}), users can select the number of inference replicas to be deployed. This exploits the consumer group feature of Apache Kafka, thereby enabling load balancing and fault-tolerance for inference. Moreover, all the interactions are done through Apache Kafka, and users have to configure the input topic (for values to predict on) and output topic (for predictions).

\subsection{Ingesting the deployed trained models with stream data for inference}
Finally, the ML/AI pipeline concludes once the trained model is ready and deployed to make predictions and recommendations through data streams. In this case, no control messages have to be sent since the input and output topics, and the input format and configuration have been previously defined in the Web-UI (Fig. \ref{fig:deploy-inference}). Users and systems just need to send encoded data streams with the data format defined to the input topic, and inference results will be immediately sent once model predictions to the output topic configured.

\section{Kafka-ML architecture}
\label{sec:architecture}

\begin{figure}[h]
    \centering
    \includegraphics [scale=0.35]{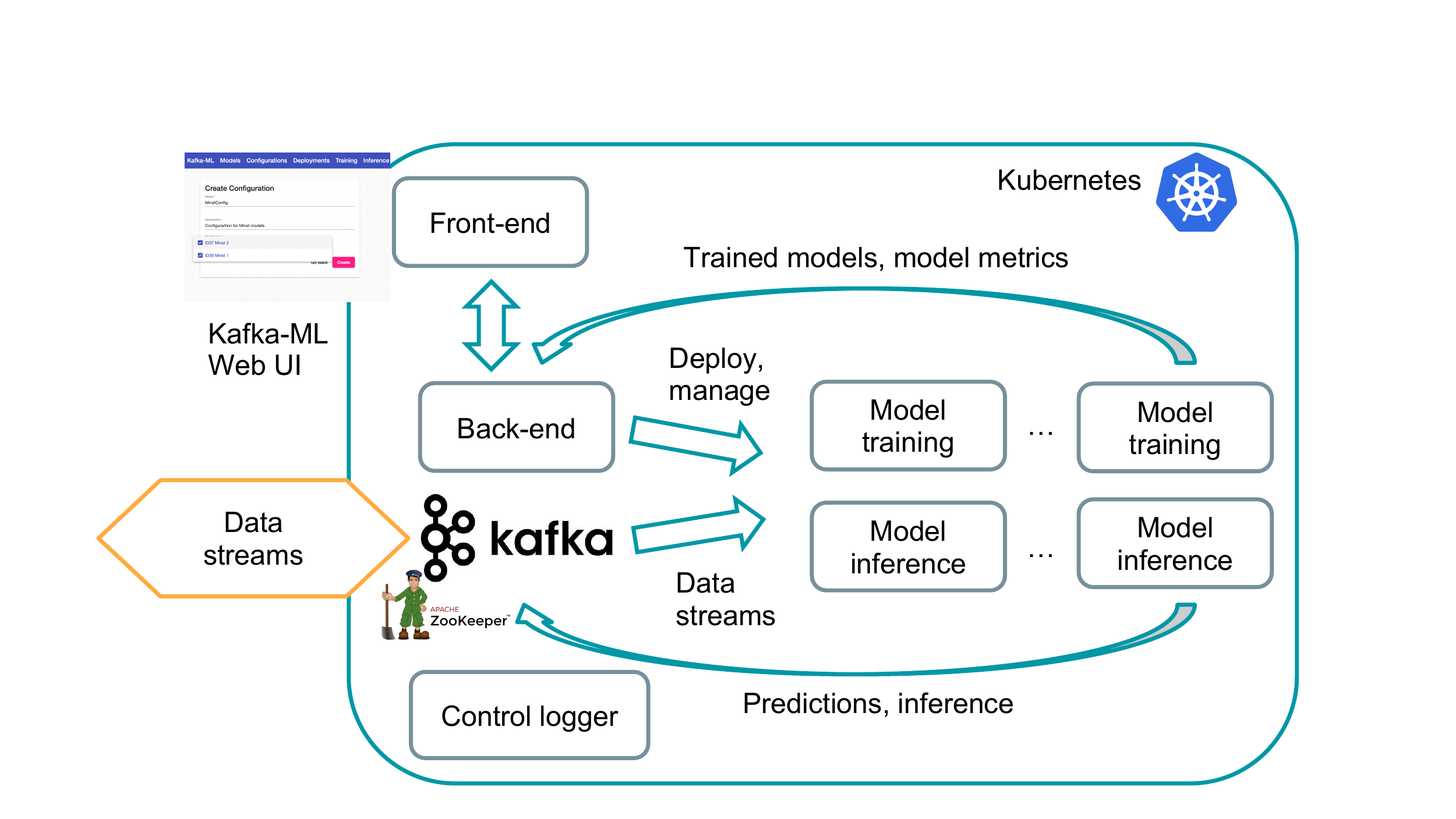}
    \caption{Overview of Kafka-ML architecture}
    \label{fig:architecture}
\end{figure}

The Kafka-ML architecture comprises a set of components based on the single-responsibility principle, comprising a microservice architecture. All of these components have been containerized so that they can run as Docker containers. This does not only enable easy portability of the architecture, isolation between instances and fast setup support for different platforms but also their management and monitoring through a container orchestration platform like Kubernetes \cite{kubernetes}. Kubernetes enables continuous monitoring of containers and their replicas to ensure that they continuously match the status defined for them, in addition to allowing other features for production environments such as high availability and load balancing. Kubernetes manages the life cycle of Kafka-ML and its components. Kafka-ML is an open-source project and its implementation, configurations, Kubernetes deployment files and some examples can be found in our GitHub repository\footnote{https://github.com/ertis-research/kafka-ml}. An overview of the Kafka-ML architecture is shown in Fig. \ref{fig:architecture}, and below each component is detailed.

\subsection{Front-end}
The front-end provides a management Web UI where users can perform the operations available in Kafka-ML such as the creation of ML models and configurations and the deployment of them for training and inference in a user-friendly and accessible way. The front-end makes use of the RESTful API offered by the back-end, and it has been implemented using the popular TypeScript framework for Web development Angular. Since the front-end and back-end have been clearly differentiated, this architecture opens the door to the integration of third-party applications and the creation of new front-end (e.g., a smartphone application).

\subsection{Back-end}
The back-end component serves a RESTful API to manage all the information contained in Kafka-ML such as ML models, configurations and deployments. This component is in contact with the corresponding Kubernetes API to deploy and manage training and inference of configurations and ML models when ordered by users. Moreover, the back-end also receives the trained ML models and metrics after training a configuration. These trained models can be downloaded or deployed for later inference. This component has been implemented through the Python Web framework Django along with the official Python client library for Kubernetes\footnote{https://github.com/kubernetes-client/python} for the deployment and management of Kubernetes components. 

\subsection{Model Training}
Once the back-end deploys a configuration, a Job, a deployable unit in Kubernetes will be executed per Kafka-ML model for training and containerizing a Docker container. Algorithm \ref{algorithm:training} describes the procedure of the training Job. Note that some steps such as management of exceptions and data stream decoding have not been included for simplicity. Firstly, the Job downloads the ML model from the back-end. Next, it starts receiving the control stream until it receives the control stream expected, i.e., it matches the deployment\_id received. Training and optionally evaluation are performed through the data stream received in the control stream message. Finally, the Job submits the trained model and the training and evaluation metrics to the back-end.  

\begin{algorithm}
 \KwData{model\_url, training\_kwargs, evaluation\_kwargs, deployment\_id, stream data}
 \KwResult{Trained ML model and training and evaluation metrics}
 model $\leftarrow$ downloadModelFromBackend(model\_url)\;
 trained $\leftarrow$ False\;
 \While{not trained}{
  msg $\leftarrow$ readControlStreams()\;
  \If{deployment\_id == msg.deployment\_id}{
   training\_stream $\leftarrow$ readStream(msg.topic)\;
   \If{msg.validatition\_rate $>0$}{
       training\_stream  $\leftarrow$ take(data\_stream,       msg.validation\_rate)\;
        evaluation\_stream $\leftarrow$ split(data\_stream, msg.validation\_rate)
    }
    training\_res $\leftarrow$ trainModel(model, training\_kwargs, training\_stream)\;
    \If{msg.validatition\_rate $>0$}{
    evaluation\_res $\leftarrow$ evaluateModel(model, evaluation\_kwargs, evaluation\_stream)\;
    }
    uploadTrainedModelAndMetrics(model\_url, model, training\_res, evaluation\_res)\;
    trained $\leftarrow$ True\;
    }
   }
 \caption{Training algorithm in Kafka-ML}
 \label{algorithm:training}
\end{algorithm}

\subsection{Model Inference}
After training an ML model and deploying it for inference through the back-end, a Replication Controller (a Kubernetes component that ensures that a specified number of replicas are running at all times) with the replicas established will be executed in Kubernetes along with its corresponding Docker containers. Algorithm \ref{algorithm:inference} describes the procedure of the inference in Kafka-ML. Once downloaded the trained model from the back-end, this component starts receiving stream data to then make a prediction with the stream received and sending it through the Kafka output topic configured. When having multiple Kafka brokers and partitions, the Replication Controller exploits the consumer group feature of Apache Kafka by matching replicas and partitions to provide load balancing and higher data ingestion.

\begin{algorithm}
 \KwData{model\_url, input\_topic, output\_topic, input\_configuration, stream data}
 \KwResult{Predictions to Apache Kafka}
 model $\leftarrow$ downloadTrainedModelFromBackend(model\_url)\;
 deserializer $\leftarrow$ getDeserializer(input\_configuration)\;
 \While{True}{
  stream $\leftarrow$ readStreams(input\_topic)\;
  data $\leftarrow$ decode(deserializer, stream)\;
  predictions $\leftarrow$ predict(model, data)\;
  sendToKafka(predictions, output\_topic)\;
  }
 \caption{Inference algorithm in Kafka-ML}
 \label{algorithm:inference} 
\end{algorithm}

\subsection{Control logger}

The control logger component is just responsible to consume the control messages received in Kafka-ML, and send them to the back-end. These control messages are used for two purposes in the back-end: 1) allowing to send them again to other deployed configurations without the need to send the entire training and evaluation data stream, which is possible since Apache Kafka keeps the data streams in the distributed log; and 2) auto-configuring the inference input format and configuration with the information received in the control messages. The input format and configuration are not directly configured in the Kafka-ML Web UI but they are defined in the control messages, so this facilitates the work for users in defining the data parameters when deploying an ML model for inference. During training, Jobs receive the control data stream and thereby this information.

\subsection{Apache Kafka and ZooKeeper}
To facilitate the deployment and management of Apache Kafka, and also to leverage the possibilities offered by Kubernetes, we have deployed Apache Kafka\footnote{https://github.com/wurstmeister/kafka-docker} and Apache ZooKeeper\footnote{https://hub.docker.com/\_/zookeeper} (required by Apache Kafka for synchronization) as Jobs using Docker containers in Kubernetes. We have enabled their exposure through a Kubernetes service both internally to the rest of the components and externally to enable other systems to send the data streams.

\section{Data stream management through Apache Kafka distributed log}
\label{sec:storage-kafka}

\begin{figure*}[h]
    \centering
    \includegraphics [scale=0.42]{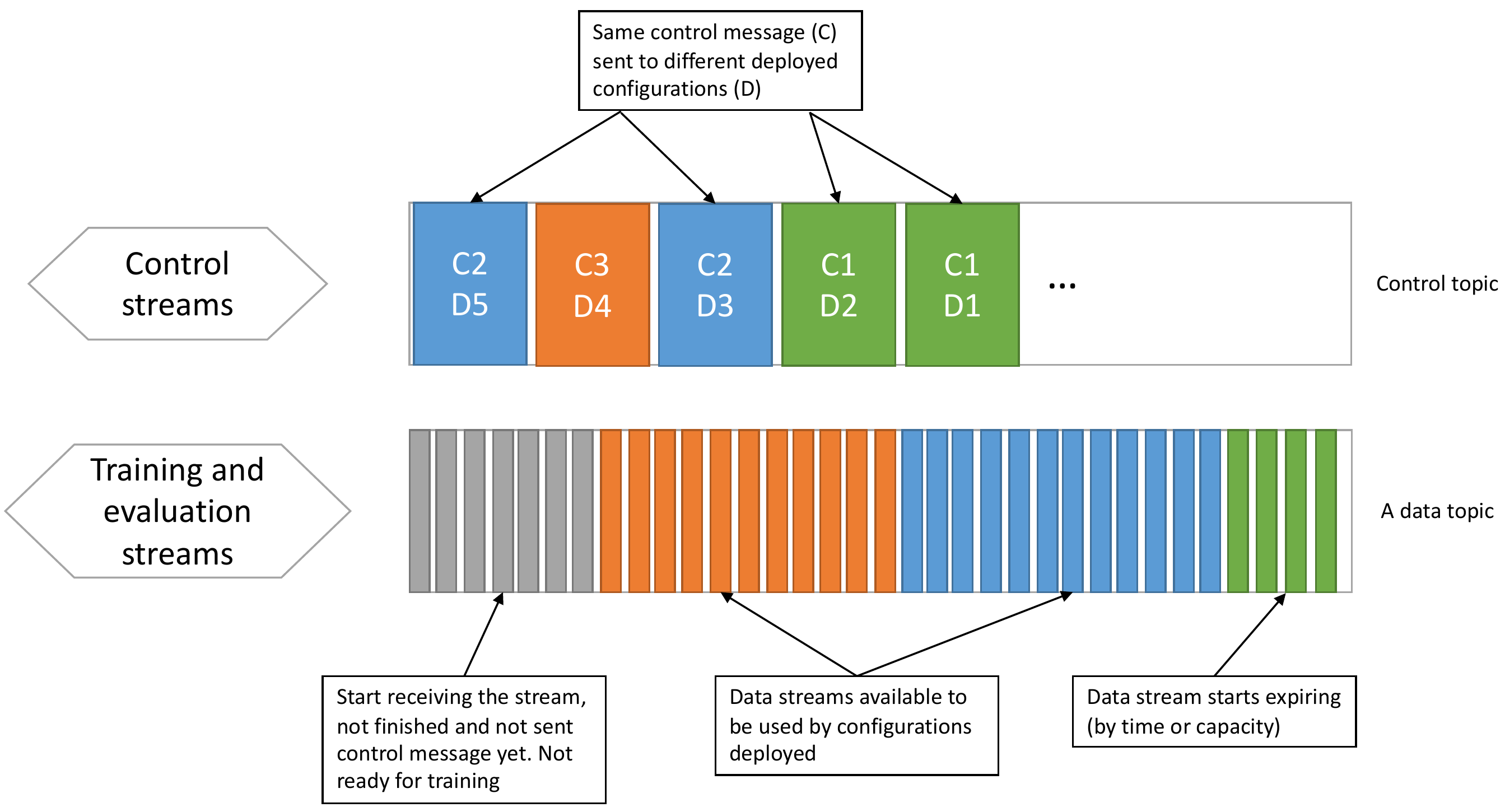}
    \caption{Data stream management in Kafka-ML}
    \label{fig:kafka-storage}
\end{figure*}

As discussed in Section \ref{sec:background}, the distributed log provided in Apache Kafka enables consumers to move along the log and read data streams as they wish. This is useful when a component/system that has to process all data at once (e.g., a training Job) fails and has to recover all data stream. In traditional message queue systems where each message may be deleted after consumption, a datastore may be needed to ensure there is no data loss in these situations. 

On the other hand, since data streams can be configured to be kept in the log, these streams can be reused for training into other deployed configurations and ML models without the need to send the whole data stream again. The only requirement is to send the corresponding control message (tens of bytes) to the desired deployed configuration as long as the data stream is available in Apache Kafka with the retention policy established. An example of this functionality is shown in Fig. \ref{fig:kafka-storage}. Firstly, the first data stream (green data) was sent along with its control message (C1) to the deployed configuration D1. A control message C1 was sent again to allow configuration D2 to consume the same data stream. In the current distributed log state, this data stream is expiring and cannot be longer reused to another deployed configuration. The data stream associated with the control message C2 has been sent to the deployed configurations D3 and D5, which can still be reused for new configurations that want to use this stream data again. Finally, the data stream on the left in the Training and evaluation stream is now entering the distributed log and the control message has not been sent yet since the data stream has not finished.

To allow training and evaluation tasks to freely move along the stream data, control messages do not only specify the topic where the data stream are but also what it is their position in the distribution log. This follows the following list format provided by the KafkaDataset connector from TensorFlow/IO: [topic:partition:offset:length]. For instance, the example [kafka-ml:0:0:70000] specifies that in the topic kafka-ml  and its partition 0, the data stream is available from the offset position 0 to 70000. Through this control message, Kafka-ML informs deployed configurations where exactly the data streams are. In the Kafka-ML Web UI, a form is available where users can see the list of the data stream that has been sent to Kafka-ML, which can be reused for other configurations.

As discussed, this behavior depends on the retention policy established in Apache Kafka. Current retention strategies within the Apache Kafka \textit{delete} retention policy are:

\begin{enumerate}
    \item Retention bytes: Control the maximum size a partition can grow to before Kafka will discard old log segments to free up space. Default not applicable.
    \item Retention ms: Control the maximum time a log will be retained before old log segments will be discarded to free up space. Default to 7 days.
\end{enumerate}

Note that Apache Kafka provides another retention policy known as the \textit{compact} policy, which ensures that Kafka will retain at least the last known value for each message key for a single topic partition. Nevertheless, due to the necessity of neither loss nor compacting data, the delete retention policy would be preferred for Kafka-ML instead.

\section{Validation}
\label{sec:validation}

\begin{figure}[h]
    \centering
    \includegraphics [scale=0.5]{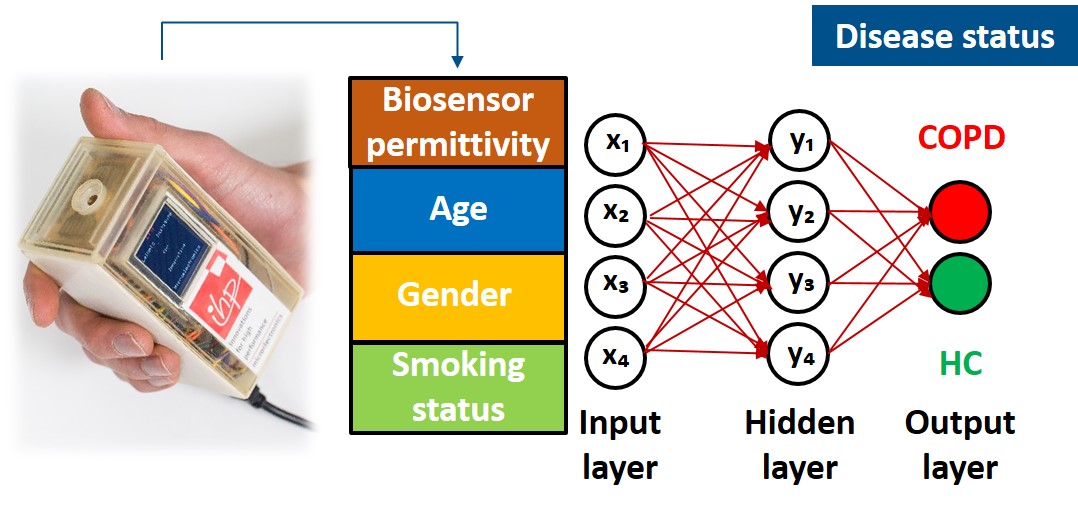}
    \caption{Architecture of a neural network model used for the classification of COPD and HC samples available at the Exasens dataset, using IHP's permittivity biosensor}
    \label{fig:biosensor}
\end{figure}

The open access Exasens dataset, available at the UCI machine learning repository\footnote{https://archive.ics.uci.edu/ml/datasets/Exasens}, was used in this study for the validation of Kafka-ML through classification of saliva samples of Chronic Obstructive Pulmonary Disease (COPD) patients and Healthy Controls (HC). This novel dataset contains information on saliva samples collected from groups of respiratory patients including COPD and HC \cite{7t0z-pd65-20}. The attributes of the dataset, used for the classification of subjects, include demographic information of patients (age, gender, and smoking status) as well as the dielectric properties of the characterized saliva samples for every class, as shown in Fig. \ref{fig:biosensor} \cite{soltani2019development}. For computational purposes, the non-quantitative attributes—diagnosis, gender, and smoking status—were converted into numerical values using the following labels: diagnosis (COPD (1)–HC (0)), gender (male (1)–female (0)), smoking status (smoker(3)–ex-smoker (2)–non-smoker (1)). The metrics, models, and sections of the data used for analytics in this work are available at GitHub\footnote{https://github.com/Pouya-SZ/HCOPD}. To improve the performance of the ML classifier, the datasets was standardized to the Gaussian standard normal distribution with zero mean and unit variance. In addition, the 5-fold cross-validation method was implemented for the evaluation of the model in this work, thus preventing overfitting and providing reliable results. Therefore, for every cross-validation fold, the dataset was split into test--train subsets with the ratio of 20—80\%, respectively. The data preparations and ML implementations were performed on the JupyterLab environment using Keras 2.2.5 and Scikit-learn 0.22 libraries of Python.

An Avro encoding format was designed for both the training and label data since the input data has multiple variables. The source code of this validation is included in the examples of Kafka-ML\footnote{https://github.com/ertis-research/kafka-ml/tree/master/examples/ HCOPD\_Avro\_format}. The Keras model source code was inserted as defined at the HCOPD GitHub repository (Listing 2).

\begin{lstlisting}[caption={HCOPD Keras ML code used in Kafka-ML}, label={list:copd}]
model = tf.keras.Sequential([
tf.keras.layers.Dropout(0.2, 
    input_shape=(4,)),  
tf.keras.layers.Dense(4, 
    activation='sigmoid'),
tf.keras.layers.Dense(2, 
    activation='softmax')
])
model.compile(tf.keras.optimizers.Adam
(lr=.0001), 
loss='sparse_categorical_crossentropy', 
metrics=['accuracy'])
\end{lstlisting}   

Through this example, we have measured and evaluated the response time of Kafka-ML regarding data stream and containerization. Latency response has been measured to study the impact of training and inference in the following cases: 1) without the Kafka integration (no data streams); 2) with the data stream integration; 3) and with both the data stream integration and containerization. Note that the training response includes the data stream ingestion and the inference response includes the latency response between a data is sent until the prediction is received. Training has been performed with a batch size of 10 and this configuration introduced into the Kafka-ML Web-UI:
``epochs=1000, steps\_per\_epoch=22, shuffle=True, verbose=0''. 

The validation was performed on a single Kubernetes cluster running on a MacBook Pro laptop with 16GB. Latency response of the Exasens dataset regarding training and inference are shown in Tables \ref{table:training} and \ref{table:inference} respectively. In the case of data streams, the latency response can be admissible taking into account the advantages seen for the ML/AI pipeline. In the case of containerization, the latency is a little higher than data streams, especially for training. For Inference is lower since Kafka is deployed in Kubernetes and thereby the network delay is smaller. We will study how to improve containerization training through distribution and GPU support.

\begin{table}[h]
\small
\centering
\caption{Training latency response (s)}
\label{table:training}
\begin{tabular}{p{1.5cm}p{1.5cm}p{2.5cm}}
\hline
Normal & Data streams & Data streams \& containerization  \\ \hline
27,37 & 29,61 &  31,44 \\
\end{tabular}
\end{table}

\begin{table}[h]
\small
\centering
\caption{Inference latency response (s)}
\label{table:inference}
\begin{tabular}{p{1.5cm}p{1.5cm}p{2.5cm}}
\hline
Normal & Data streams & Data streams \& containerization  \\ \hline
0,079 & 0,374 &  0,335  \\
\end{tabular}
\end{table}

As performed with the Exasens dataset and a classic ML model, Kafka-ML can be used to manage the ML/AI pipeline of other ML works, facilitating their evaluation and data stream management, with a publicly available GitHub source code.

\section{Related Work}
\label{sec:related-work}

To the best of our knowledge, Kafka-ML is the first framework to provide an ML/AI pipeline solution to integrate machine learning and data streams, TensorFlow and Apache Kafka. Nevertheless, other approaches have similar goals or have provided some of the functionalities offered by Kafka-ML as described below.  

NVIDIA Deep Learning GPU Training System (DIGITS) \cite{yeager2015digits} provides an interactive Web UI for training and inference of deep neural networks (DNNs) on multi-GPU systems. Unlike Kafka-ML, DIGITS is not a framework itself, but it is a wrapper for NVCaffe, Torch and TensorFlow, which provides a Web interface to those frameworks rather than dealing with them directly on the command-line. The main advantages of DIGITS are its native support for GPUs and three ML frameworks, the release of pre-trained models and the functionality to see the accuracy and loss in real-time. Nevertheless, DIGITS does not support training and inference through data streams (datasets have to be imported instead) and the deployment of these tasks through containers for scaling, it has a dependency on GPUs and may require writing a source code on top of these frameworks.

\begin{figure}[h]
    \centering
    \includegraphics [scale=0.30]{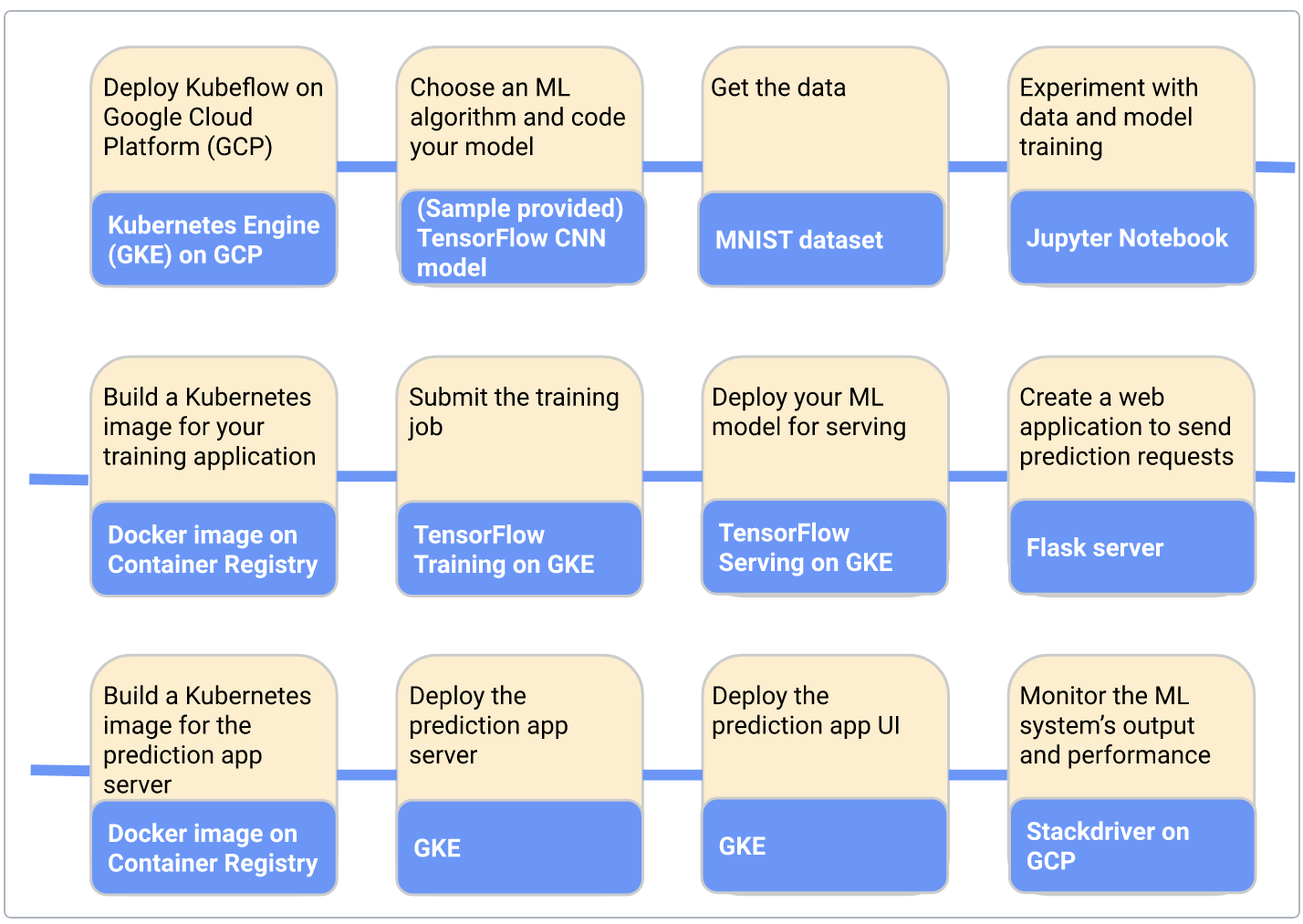}
    \caption{Kubeflow pipeline example for GCP. Source \url{https://www.kubeflow.org/docs/gke/gcp-e2e/}}
    \label{fig:kubeflow}
\end{figure}

Kubeflow \cite{kubeflow} is a powerful ML toolkit for Kubernetes. In Kubeflow, users can configure multiple steps of an ML/AI pipeline such as hyper-parameters, pre-processing, training and inference. However, when running a Kubeflow pipeline such as the official example for the Google Cloud Platform (Fig. \ref{fig:kubeflow}), there may be some steps that are not required in the Kafka-ML pipeline (Fig. \ref{fig:pipeline}), especially the ones that require to build containers for training and inference. In Kafka-ML, users merely need to interact with the Web UI for training and inference. In addition, data stream support would have to be manually developed by Kubeflow ML developers and users. In Kafka-ML, the data stream management through Apache Kafka is supported in all the pipeline. In any case, Kubeflow provides great support for Kubernetes and ML multi-frameworks, and it is supported by a large ecosystem and community that is far beyond the scope and functionalities offered by Kafka-ML, therefore it may worth studying the way of integrating both systems in the near future.

MOA \cite{DBLP:journals/jmlr/BifetHKP10} is a framework for online learning and data stream mining. MOA provides a graphical interface where users can execute and visualize ML tasks, including a collection of ML algorithms implementations for classification, regression and clustering among others.  Although Kafka-ML supports data streams, Kafka-ML and TensorFlow are not well supporting (yet) online learning. On the other hand, Kafka-ML provides support for TensorFlow/Keras models, and its large community, instead of creating a new framework with own source code that could limit its adoption. Scikit-multiflow  \cite{skmultiflow} is another framework for online learning, in this case for the popular framework scikit-learn, however it does not provide a Web interface nor a full control of an ML/AI pipeline.

Kafka-ML follows a different approach than other distributed data stream frameworks such as Apache SAMOA \cite{samoa}, Apache Flink \cite{carbone2015apache} and the Lambda architecture \cite{Diaz2016}, \cite{ge2019scalable}. Apache SAMOA is currently undergoing incubation at Apache and aims to enable the development of ML algorithms through data streams without directly dealing with the complexity of underlying processing engines (e.g., Apache Storm and Apache Samza). Apache Flink provides a framework to perform computation over data streams at in-memory speed at any scale. And the Lambda architecture allows the processing of large amounts of data in real-time by having real-time and batch layers of processing. In general, these frameworks provide distributed engines for distributing any kind of computation with data streams, but they have limited support or do not have a special focus on facilitating ML/AI pipelines and popular ML/AI frameworks such as TensorFlow, and their large range of ML/AI solutions and community, as Kafka-ML does. Moreover, Kafka-ML can also enable the deployment of high availability and fault-tolerant ML/AI pipelines.

Finally, Kafka-ML is related to some extent to AutoML projects such OpenML \cite{openML} and Google Cloud AutoML \cite{googleAutoML}. OpenML is a web platform where users can openly share, upload and explore results, scientific tasks, data analysis flows and datasets. Results and metrics of ML models can also be shared and compared (using configurations) in Kafka-ML. Moreover, data streams can also be managed and shared as seen in Section \ref{sec:storage-kafka}. Google Cloud AutoML provides high-quality ML models with little effort and no advanced knowledge of the subject. Reaching the quality of these models is beyond the scope of Kafka-ML, however, Kafka-ML provides an accessible and user-friendly platform, where only a few lines of ML model source code are required to start an ML/AI pipeline. Furthermore, Kafka-ML is an open-source project available for both experts and non-experts on ML/AI.

\section{Conclusions and Future work}
\label{sec:conclusions-future}
In this paper, Kafka-ML, an open-source framework to manage the pipeline of ML/AI applications through data streams has been presented. Kafka-ML exploits the popular data stream system Apache Kafka and the Python ML framework TensorFlow to integrate both ML/AI and data streams. Kafka-ML is characterized by its accessibility and easy-use since with only a few lines source code, users can create an ML model in its Web UI to control the ML/AI pipeline, creating configurations to evaluate different ML models, training, validating and deploying trained models for inference. Moreover, a novel approach based on the distributed log of Apache Kafka has been adopted to have full control of the data streams received in Kafka-ML, enabling its ML/AI applications to reuse these data streams and maybe removing their dependency on data storage or file systems. Kafka-ML is fully containerized, and deployed its components (training and inference). Docker and Kubernetes are in charge of containerization and orchestrating the Kafka-ML architecture for fault-tolerance and high availability respectively. Kafka-ML is openly available in GitHub to be used and improved by both experts and non-experts on ML/AI adopting data streams.

As future work, we have pointed out the following challenges and improvements to Kafka-ML:
\begin{itemize}
    \item Distributed inference. Deep neural network layers can be partitioned into multiple and independent ML models, and through intermediary exits \cite{teerapittayanon2017distributed}, their execution can be optimized in the Fog, Edge and Cloud computing paradigms. The objective is to enable the training and partition of ML models in Kafka-ML, to later deploy them in the IoT-Cloud continuum. New architectures to support the whole data flow between layers are also required.
    
    \item Distributed training. Currently, training is performed in a single container that may not be enough for large neural networks. Other approaches for distributed training in Kubernetes such as Kubeflow and GPU support should be explored in this regard. 
    
    \item Support for other ML frameworks. This will depend on the developments of other ML frameworks to enable Apache Kafka as TensorFlow did with TensorFlow/IO. In any case, new data stream connectors to other ML frameworks can be explored. 
    
    \item IoT and ML/AI. The IoT is taking place into the ML/AI pipeline as demonstrated by initiatives such as uTensor\footnote{https://github.com/uTensor/uTensor} and TensorFlow lite\footnote{https://www.tensorflow.org/lite} for on-device inference. The generation of ML models for IoT devices and even its installation from Kafka-ML could expand the ML/AI pipeline until the IoT.
        
    \item Integrate other processing tasks. Finally, many applications such as Structural Health Monitoring may use ML/AI but also other statistical and processing tasks that may require the same data stream. Therefore, Kafka-ML could also manage these non-ML/AI tasks to integrate them with the data stream used.
    
\end{itemize}

\section*{Acknowledgments}
This work is funded by the Spanish projects RT2018-099777-B-100 (``rFOG: Improving latency and reliability of offloaded computation to the FOG for critical services'') and UMA18FEDERJA-215 (``Advanced Monitoring System based on Deep Learning Services in Fog''). Cristian Mart\'in is with a postdoc grant from the Spanish project TIC-1572 ("MIsTIca: Critical Infrastructures Monitoring based on Wireless Technologies") and his research stay at IHP has been funded through a mobility grant from the University of Malaga and IHP funding. We would like to express our anonymous gratitude to Kai Wähner for his inspiration and ideas through numerous articles and GitHub repositories on Kafka and its combination with TensorFlow. 

\bibliographystyle{IEEEtran}
\bibliography{IEEEabrv,template}

\end{document}